\pdfoutput=1
\documentclass[11pt]{article}

\usepackage[]{EMNLP2022}

\usepackage{times}
\usepackage{latexsym}

\usepackage[T1]{fontenc}
\usepackage[utf8]{inputenc}

\usepackage{microtype}

\usepackage{inconsolata}

\usepackage{graphicx}
\usepackage{multirow}
\title{ViQA-COVID: COVID-19 Machine Reading Comprehension Dataset for Vietnamese}

\author{Hai-Chung Nguyen-Phung \,
Ngoc C. Lê \,
Van-Chien Nguyen\\
\textbf{Hang Thi Nguyen} \,
\textbf{Thuy Phuong Thi Nguyen}
\\
Hanoi University of Science and Technology}
\begin{document}
\maketitle
\begin{abstract}
After two years of appearance, COVID-19 has negatively affected people and normal life around the world. As in May 2022, there are more than 522 million cases and six million deaths worldwide (including nearly ten million cases and over forty-three thousand deaths in Vietnam). Economy and society are both severely affected. The variant of COVID-19, Omicron, has broken disease prevention measures of countries and rapidly increased number of infections. Resources overloading in treatment and epidemics prevention is happening all over the world. It can be seen that, application of artificial intelligence (AI) to support people at this time is extremely necessary. There have been many studies applying AI to prevent COVID-19 which are extremely useful, and studies on machine reading comprehension (MRC) are also in it. Realizing that, we created the first MRC dataset about COVID-19 for Vietnamese: ViQA-COVID and can be used to build models and systems, contributing to disease prevention. Besides, ViQA-COVID is also the first multi-span extraction MRC dataset for Vietnamese, we hope that it can contribute to promoting MRC studies in Vietnamese and multilingual. We will publicly release ViQA-COVID soon.
\end{abstract}
\section{Introduction}
Omicron - a dangerous variant of SARS-CoV-2 has shown its danger in recent months. Specifically, on average, each day there are around five hundreds thousands new cases and around ten thousands deaths worldwide. The uncontrollably rapid spread leads to the overwhelming of resources in disease prevention: medical staff, medical equipment manufacturing workers, data analysts, anti-epidemic support teams, etc. In the long run, this will have serious economic, social, as well as human impacts.

% Therefore, the application of intelligent and automated systems such as artificial intelligence and machine learning models to assist and replace work for humans is essential: to help reduce the work load on staffs and hence, can help the fight against the pandemic more effective. Many researches on artificial intelligence have been applied and brought positive effects, greatly benefiting society. Among them, natural language processing (NLP) systems have worked extremely well and there will be many more applications that can be deployed in the future. Specifically, NLP applications in COVID-19 prevention can be mentioned as: information extraction; patient and location information query; automatic documents summarization; automatic question and answer about the pandenmic system; epidemic related sentiment analysis, etc.

In Vietnam, the number of cases is increasing very quickly. The information of the cases must be updated continuously to support the medical team to capture information and promptly treat the patient. On national portals, important information about the epidemic such as the number of cases, time and location related to the epidemic, people in contact with the patient, also needs to be updated quickly so that people can grasp the information and protect themselves. In addition, the hotlines and portals receive a lot of questions and reflections from the people every day. It can be seen that the amount of data generated daily is very large and difficult to handle manually. Thus, a system to extract information and answer questions like the machine reading comprehension (MRC) system is extremely necessary at the present time. It will be an aid for the prevention of COVID-19 or even other diseases in the future.

% In the above application systems, MRC systems are used in many applications, but most of them only work in high-resources languages such as English, Chinese. It can be seen that dataset plays a very important role in the development of NLP systems in particular and machine learning in general. Thus, low-resource languages enrichment is the first step to create a MRC system for these languages. With that in mind, we introduce ViQA-COVID, a multi-span extraction MRC dataset related to COVID-19 for Vietnamese with the desire to develop Vietnamese and multilingual MRC systems and help in fighting against COVID-19. Dataset comprises question-answer pairs based on CDC case reports, assigned to COVID-19 prevention teams and news on Vietnam's reputation online newspapers.

To build a COVID-19 MRC system, a COVID-19 MRC dataset is required. As a matter of fact, sufficient MRC dataset on COVID-19 for Vietnamese has yet to be released. Therefore, we created ViQA-COVID, a multi-span extraction MRC dataset about COVID-19 for Vietnamese based on official data from Centers for Disease Control and Prevention (CDC) Vietnam and reputable online news sites. In addition, ViQA-COVID is also the first multi-span extraction MRC dataset for Vietnamese. The goal of this research is to contribute to building data sources for low-resource languages like Vietnamese.

In the next section, related works will be covered. Section \ref{secDataset} presents about datasets, statistics and annotation process. Section \ref{secExp} is devoted for experiments set up. The results and benchmark are described in Section \ref{secResult}. Section \ref{secConclusion} summarizes the study and presents further research directions.
\section{Related Work}
\label{secRelated}
In recent years, COVID-19 has spurred research in many fields especially in AI related ones. In the field of computer vision, researchers \cite{Wang2020} designed COVID-Net to detect COVID-19 cases from chest X-ray (CXR) images and introduced COVIDx, a dataset consisting of 13,975 CXR images across 13,870 patient cases. In \cite{DBLP-2003-09093}, three masked face datasets: Masked Face Detection Dataset (MFDD), Real-world Masked Face Recognition Dataset (RMFRD), and Simulated Masked Face Recognition Dataset (SMFRD) that helped a lot in detecting and reminding people to wear masks (one of the most effective measures to prevent covid-19), are introduced. The image editing approach and datasets: Correctly Masked Face Dataset (CMFD), Incorrectly Masked Face Dataset (IMFD), as well as their combination - masked face detection (MaskedFace-Net) are introduced in \cite{cabani}. MaskedFace-Net has been applied to detect whether people are wearing masks and wearing them correctly.

In the field of NLP, COVID-QA \cite{moller-etal-2020-covid} is an MRC dataset consisting of 2,019 pairs of questions - answers labeled by experts, with data sources collected from CORD-19. COVID-QA is widely used in evaluating MRC tasks and applied to tasks related to COVID-19. CovidQA \cite{DBLP-2004-11339} is one of the first Question Answering datasets, consisting of pairs of questions - articles and answers that are articles related to the question. CovidDialog \cite{ju2020CovidDialog} provides a dataset including doctor-patient conversations (603 consultations and 1,232 utterances in English and 399 consultations and 8,440 utterances in Chinese). Using CovidDialog, researchers \cite{zeng2020coviddialogmodel} have developed a medical dialogue system to provide information related to the pandemic. \cite{zhang2021cough} publicly released COUGH, a COVID-19 FAQ dataset includes 15,919 FAQ items, 1,236 human-paraphrased user queries and each query has 32 human-annotated FAQ items. Phoner\_COVID \cite{PhoNER_COVID19}, \cite{le2025nested}, \cite{Le2023} are Vietnamese NER dataset about COVID-19 which defined various entities related to COVID-19 patients information. In addition, there are many research works that have been highly applicable and have greatly supported countries in preventing COVID-19.

With the rapid development of NLP in Vietnam, many new datasets have been introduced. From collecting 174 articles on the Vietnamese Wiki and through a five-phase annotate process, UIT-ViQuAD \cite{nguyen-etal-2020-vietnamese} was created with more than 23,000 question-answer pairs based on 5,109 passages. UIT-ViQuAD is a single span-extraction MRC datasets widely used in span-extraction MRC task Vietnamese besides UIT-ViNewsQA \cite{DBLP-2006-11138}, a dataset in healthcare domain consisting of 22,057 question-answer pairs based on 4,416 articles health report. In addition, ViMMRC \cite{9247161} is a multiple-choice dataset and includes 2,783 multiple-choice questions based on 417 Vietnamese texts. With the task of sentence extraction-based MRC, UIT-ViWikiQA \cite{DBLP-2105-09043} is the first Vietnamese sentence extraction-based MRC dataset, created from converting the UIT-ViQuAD dataset. UIT-ViWikiQA includes 23,074 question-answer pairs, based on 5,109 passages.

In addition to the studies on COVID-19 and MRC datasets for Vietnamese, we also consulted other famous MRC datasets such as: SQuAD1.1 \cite{rajpurkar-etal-2016-squad}, SQuAD2.0 \cite{rajpurkar-etal-2018-know}, GLUE \cite{wang-etal-2018-glue}, SuperGLUE \cite{NEURIPS2019_4496bf24}, MASH-QA \cite{zhu-etal-2020-question}, QUOREF \cite{dasigi-etal-2019-quoref} and DROP \cite{dua-etal-2019-drop}.

The above studies helped us to complete our research.
\section{Dataset}
\label{secDataset}
% \begin{figure*}[t]
% \centering
% \includegraphics[width=1\textwidth]{fig1.drawio.pdf} 
% \caption{An example include passage, question and answer from ViQA-COVID. Bold words in passage are answers}
% \label{fig1}
% \end{figure*}
In this section, ViQA-COVID, annotation processing and statistics about the dataset is described in detail. 

CDC daily receives a large amount of data on cases, reflections and questions from people. Extracting and processing information from this data source is essential to helping medical teams understand the situation and make decisions to prevent COVID-19. However, handling huge amounts of data by hand is extremely complex. In addition, unfixed-form data and complexity of Vietnamese make it difficult to handle with rule-based approach. Based on previous studies as \cite{nguyen-etal-2020-vietnamese}, \cite{zhu-etal-2020-question}, \cite{segal-etal-2020-simple}, etc., it can be seen that a MRC system based on deep learning can solve the above problems. For example: From the patient's epidemiological information, the medical team asks: "\textit{Who has the COVID-19 patient been in contact with?}". MRC system can answer correctly and the medical team can isolate and treat those people quickly. In addition, MRC system can help answer people's questions about disease, policies, ways to prevent COVID-19 and so on. To be able to achieve the aforementioned purposes, MRC system needs to train with MRC datasets. Therefore, ViQA-COVID has been created as training data for such system. 
% Figure \ref{fig1} shows an example from ViQA-COVID.
\subsection{Annotation}
The annotation team consists of three data analyst from CDC annotating and reviewing data, and two experts from CDC advising on the questions and information to annotate on the data. In general, the annotation process includes following phases:
\begin{itemize}
    \item \textbf{Collect and clean passage data from CDC:} CDC data was collected from Portal of Vietnam's Ministry of Health about COVID-19\footnote{\url{https://covid19.gov.vn}}. With limited time and resource, annotating
    all the data is not possible. Therefore, report cases are chosen on the basis of informativeness and structural diversity. Data was encrypted sensitive information (so as not to violate privacy issues), corrected typing and grammar errors. After data cleaning, a total of 537 passages were collected.
    \item \textbf{Create and cross-check question-answer pairs:} Data is manually annotated. Question-answer pairs in ViQA-COVID are based on the information CDC needs to support patients and prevent diseases, as well as questions from people about the epidemic situation. For example: "\textit{What places have patients been to?}", "\textit{Where are the epidemic locations that I need to be aware of?}", etc. Annotators will read each passage, create questions and mark spans for corresponding answers (a answer can include multi-span). Questions are diversified and avoid duplication. Question-answer pairs are cross-checked to eliminate errors.
    \item \textbf{Collect data from other sources, annotate and cross-check:} More data from reputable online portals and online news sites were collected to diversify dataset. This data is also reviewed, manually annotated and cross-checked.
    \item \textbf{Review and recheck:} To ensure data was clean and did not violate privacy issues, we reviewed and cross-checked again to complete ViQA-COVID dataset.
\end{itemize}
\begin{table*}
\centering
\begin{tabular}{l r r r}
\hline
 & Train & Dev. & Test\\
\hline
Number of passages & 284 & 139 & 114\\
Number of questions & 3408 & 1668 & 1368\\
\hline
Average passage length & 336.8 & 269.1 & 252.7\\
Average question length & 11.2 & 9.5 & 11.1\\
\hline
Passage vocabulary size & 6659 & 3882 & 3089\\
Question vocabulary size & 1071 & 606 & 601\\
\hline
Number of multi-span answers (\%) & 712 (20.9) & 351 (21.0) & 291 (21.3)\\
Number of single-span answers (\%) & 2288 (67.1) & 1147 (68.8) & 927 (67.8)\\
Number of non-span answer (\%) & 408 (12.0) & 170 (10.2) & 150 (10.9)\\
\hline
\end{tabular}
\caption{ViQA-COVID overview}
\label{table2}
\end{table*}
\begin{table}
\centering
\begin{tabular}{l l}
\hline
Question Types & Question Words\\
\hline
What (19.3\%) & l\`{a} g\`{i} (10.5\%)\\
Where (17.2\%) & \dj\^{a}u (7.3\%)\\
When (36.6\%) & ng\`{a}y n\`{a}o (10.4\%)\\
Who (8.4\%) & ai (6.2\%)\\
How (3.3\%) & th\'{\^{e}} n\`{a}o (1.1\%)\\
How many (10.2\%) & bao nhi\^{e}u (9.6\%)\\
\hline
\end{tabular}
\caption{Question types and questions words distribution in ViQA-COVID}
\label{table1}
\end{table}
\subsection{Statistics}
% \begin{figure}[t]
% \centering
% \includegraphics[width=0.45\textwidth]{fig2.drawio.pdf} 
% \caption{Question types and questions words distribution in ViQA-COVID}
% \label{fig2}
% \end{figure}
ViQA-COVID after completion has a total of 6,444 question-answer pairs based on 537 passages. To our knowledge, ViQA-COVID is the first multi-span extraction MRC dataset on COVID for Vietnamese. Details of the statistics are in Table \ref{table2}. It can be seen that, because ViQA-COVID is a domain-specific dataset (COVID-19 and Health), the vocab size is not too large. In addition, the percentage of multi-span answers is quite high compared to most multi-span MRC datasets, around 20\%.

Question types in the dataset is distributed as follows: What: 19.3\%, How: 3.33\%, How many: 10.2\%, Where: 17.16\%, When: 36.61\%, Who: 8.38\%, Others: 5.02\%. Like many others languages, each type of question may be expressed in numerous ways. Statistical description of question words in ViQA-COVID is shown in Table \ref{table1}.
\begin{figure*}[t]
\centering
\includegraphics[width=0.7\textwidth]{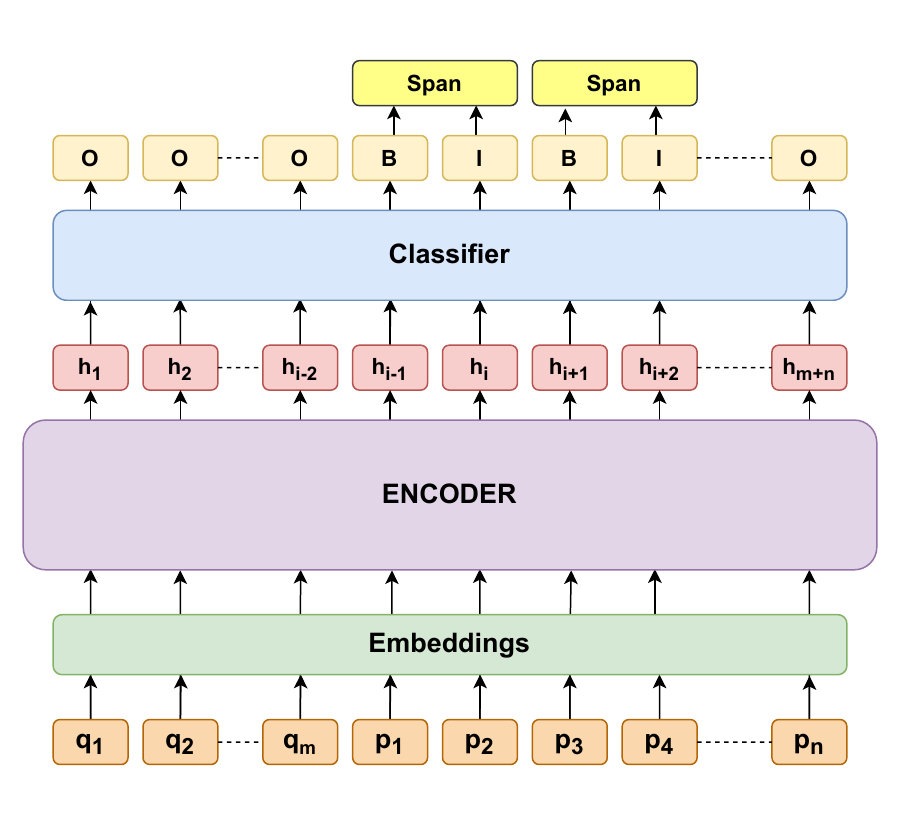} 
\caption{Illustrating the sequence tagging approach for multi-span questions. In which, $\{q_{j}\}^{m}_{j=1}$ are question tokens, $\{p_{k}\}^{n}_{k=1}$ are passage tokens and $\{h_{i}\}^{m+n}_{i=1}$ are the contextualized representations of the input tokens.}
\label{fig3}
\end{figure*}
\section{Experiments}
\label{secExp}
In this section, we present experiments with the state-of-the-art MRC models on ViQA-COVID.
\subsection{Models}
Since BERT \cite{devlin} - a pretrained model using Transformer \cite{NIPS2017_3f5ee243} architecture appeared in 2019, it has created a strong development in the field of natural language processing. State-of-the-art performance on NLP tasks increased rapidly thanks to improved models from both BERT and the Transformer architecture. It can be said that they are the two main factors that create a new era for NLP. In this experimental part, we used variants of BERT to evaluate on ViQA-COVID. These models have achieved state-of-the-art results on many MRC tasks.
\begin{itemize}
    \item \textbf{mBERT}: twelve layers with twelve self-attention heads BERT is trained on multi-lingual datasets (including Vietnamese). Since its launch in 2019, mBERT has performed very well in multi-lingual MRC and NLP tasks.
    \item \textbf{XLM-R} \cite{conneau}: based on RoBERTa \cite{roberta} - an optimal BERT-based approach, XLM-R was trained on over two terabytes of cleaned CommonCrawl \cite{wenzek2019ccnet} data in 100 languages. XLM-R outperformed mBERT in many cross-lingual benchmarks and other tasks. We evaluated two model - XLM-R$\mathrm{_{base}}$: 12 layers with 8 self-attention heads and XLM-R$\mathrm{_{large}}$: 24 layers with 16 self-attention heads.
    \item \textbf{PhoBERT} \cite{phobert}: based on RoBERTa, PhoBERT is a Vietnamese model which improved the state-of-the-art many Vietnamese NLP tasks. PhoBERT is trained on over 20 gigabytes of word-level data (while other models train with syllable data). We also evaluated two models: PhoBERT$\mathrm{_{base}}$: 12 layers with 12 self-attention heads and PhoBERT$\mathrm{_{large}}$: 24 layers with 16 self-attention heads
\end{itemize}
\begin{table}
\centering
\begin{tabular}{l r r r r}
\hline
Passage Length &  Train & Dev. & Test & Total\\
\hline
< 128 tokens        & 0 & 0 & 2 & 2\\
128 - 256 tokens    & 12 & 1 & 2 & 15 \\
256 - 384 tokens    & 25 & 11 & 8 & 44\\
384 - 512 tokens    & 38 & 20 & 18 & 76\\
$\geq$ 512 tokens   & 260 & 119 & 96 & 475\\
\hline
& 335 & 151 & 126 & 612\\
\hline
\end{tabular}
\caption{Passage length statistics}
\label{table4}
\end{table}
\subsection{Input Processing}
Statistics from Table \ref{table4} show that most passages are in excess of 512 tokens in length. Whereas maximum length of models' input feature is 512 tokens. To deal with very long passage, we split one example into input features, each of the length is shorter than model's maximum length. In case the answer lies at the position that long passage was split, we create an overlap feature between two features (controlled by stride parameter).

PhoBERT is trained with both syllable-level and word-level tokens. Unlike English, words in Vietnamese can be compound words, i.e. one word with single meaning may be a combinations of two or more single words and in most of the cases, the meaning of the compound word is very different from their components. Thus, input sentences are segmented by word segmentation which can represent them in either syllable or word-level. Therefore, word segmentation joins syllables with a "\textit{\_}" sign to indicate it's a word and makes sentences have clearer meanings. With that idea, PhoBERT outperformed XLM-R in many Vietnamese-specific NLP tasks. In our experiment, we use RDRSegmenter \cite{NguyenNVDJ2018} from VnCoreNLP \cite{vu-etal-2018-vncorenlp} as word and sentence segmentation.
\begin{table*}[ht]{
\centering
\footnotesize
\begin{tabular}{l l c c c c c c c c c c c c}
\hline
\multicolumn{2}{c}{\multirow{3}{*}{Model}} & \multicolumn{6}{c}{Dev.} & 
\multicolumn{6}{c}{Test} \\
\cline{3-14} &
{} &
\multicolumn{2}{c}{Single-Span} &
\multicolumn{2}{c}{Multi-Span} & 
\multicolumn{2}{c}{All} &
\multicolumn{2}{c}{Single-Span} &
\multicolumn{2}{c}{Multi-Span} & 
\multicolumn{2}{c}{All} \\
\multicolumn{2}{c}{}& EM & F1 & EM & F1 & EM & F1 & EM & F1 & EM & F1 & EM & F1 \\ \hline
mBERT  & & 45.10 & 51.09 & 30.52 & 61.81 & 40.83 & 54.44 & 46.28 & 51.14 & 37.64 & 65.98 & 43.49 & 55.96\\
\hline
PhoBERT$\mathrm{_{base}}$ & & 61.86 & 72.73 & 30.59 & 54.12 & 51.37 & 66.49 & 54.90 & 74.39 & 34.99 & 60.87 & 54.89 & 70.01\\

PhoBERT$\mathrm{_{large}}$ & & 62.13 & 72.48 & 32.74 & 56.70 & 52.28 & 67.19 & 64.65 & 74.21 & 37.25 & 62.14 & 55.77 & 70.30\\
\hline
XLM-R$\mathrm{_{base}}$  & & 78.90 & 83.32 & 33.20 & 71.95 & 64.62 & 79.77 & 81.23 & 85.13 & 41.27 & 77.83 & 68.34 & 82.78\\ 

XLM-R$\mathrm{_{large}}$  & & \textbf{82.74} & \textbf{86.79} & \textbf{38.20} & \textbf{75.84} & \textbf{68.82} & \textbf{83.37} & \textbf{85.11} & \textbf{89.24} & \textbf{44.44} & \textbf{79.10} & \textbf{72.00} & \textbf{85.97}\\
\hline
\end{tabular}
\caption{Performances on development set and test set}
\label{table5}
}
\end{table*}
\begin{table}
\centering
\begin{tabular}{l r r}
\hline
Question Types &  Dev. errors & Test errors\\
\hline
When & 173 & 163\\
Where & 98 & 84\\
Who & 83 & 72\\
Others & 135 & 95\\
\hline
\end{tabular}
\caption{Question type errors on development set and test set}
\label{table6}
\end{table}
\subsection{Multi-span Approach}
For the BERT-style models, we use sequence tagging approach \cite{segal-etal-2020-simple} for multi-span questions. Instead of predicting start and end probabilities like single-span questions, we predict the tag for each token. The familiar tags used are B, I, O, where B is the starting token and I is the subsequent token in output span, O is the token that is not part of an output span. Multi-span can be extracted based on B, O tokens. Figure \ref{fig3} illustrates this approach in detail.
\subsection{Training}
BERT-style models have maximum input features length of 384 (PhoBERT of 256) with stride parameter of 128. We fine-tuned models with AdamW \cite{loshchilov2018decoupled}, weight decay of 0.01, learning rate of 5e-5 and batch size of 32, in 30 training epochs on a NVIDIA Tesla P100 GPU via Google Colaboratory. Task performance was evaluated after each epoch on the development set.
\section{Results}
\label{secResult}
We evaluated models' performance on ViQA-COVID using exact match (EM) and F1-score. Results are shown in Table \ref{table5}. In which, XLM-R$\mathrm{_{large}}$ outperforms other models with 83.37\% F1-score and 68.82\% EM on development set and 85.97\% F1-score and 72.00\% EM on test set. We also evaluated the performance of the models on single-span and multi-span answers. The models are quite accurate in predicting single-span answers but still have difficulties with multi-span answers, especially in terms of exact matching. Overall, XLM-R$\mathrm{_{large}}$ performed quite well and the difficulty of ViQA-COVID is not too hard when compare to other MRC datasets.
\subsection{Error analyst}
Through empirical analysis with the best model XLM-R$\mathrm{_{large}}$, we have counted the number of incorrect answers in the development set and test set. The development set has 489/1,668 incorrect answers of which 162 multi-span, 246 single-span and 81 non-span answers. The test set has 414/1,368 incorrect answers of which 141 multi-span, 203 single-span, and 70 non-span answers. We divide these errors into four groups:
\begin{itemize}
    \item The first group consists of answers that have the correct number of spans but have an excess or lack of words. The cases are mostly long addresses or time periods (e.g. “\textit{20/5/2020 to 30/5/2020}” but the model can only predict “\textit{20/5}” or “\textit{30/5}”). These are also common mistakes in sequence tagging models.
    \item The second group includes answers that have an excess or lack of span. Mainly occurs when encountering questions about many places or about many people. For example: answering a question that lists people who have been in contact with the patient but also lists those who have not.
    \item The third group are completely incorrect answers (answers that have no correct span), often occurring in passages having a lot of noise. For example: Patient's epidemiological report contains multiple dates, including dates of admission. When answering the question about the date of admission for COVID-19 infection, the model easily mistakenly answered to the date the patient was hospitalized for another illness because of the same keyword "\textit{admission}".
    \item The fourth group includes incorrect answers on other types of questions.
\end{itemize}
The statistics of the incorrect answers are shown in Table \ref{table6}.
\section{Conclusion}
\label{secConclusion}
In this study, we introduced ViQA-COVID, the first multi-span MRC dataset about COVID-19 for Vietnamese. Our dataset consists of 6,444 question-answer pairs based on 537 passages related to COVID-19. We also experimented with different the state-of-the-art MRC models on ViQA-COVID. The results show that, XLM-R$\mathrm{_{large}}$ outperforms other models with 83.37\% F1-score and 68.82\% EM on development set and 85.97\% F1-score and 72.00\% EM on test set. We hope that our dataset will contribute to the prevention of COVID-19 as well as the development of NLP for Vietnamese and multilingual.
\renewcommand{\refname}{References}
\bibliographystyle{plain}

\end{document}